\documentstyle[]{article}

\makeatletter
\newcounter{enums}    

\def\enumsentence{\@ifnextchar[{\@enumsentence}
{\refstepcounter{enums}\@enumsentence[(\theenums)]}}
\long\def\@enumsentence[#1]#2{\begin{list}{}{
	\advance\leftmargin by 0.5em
	\advance\labelwidth by 0.5em}%
\item[#1]#2
\end{list}}

\newcounter{tempcnt}

\def\@item[#1]{\if@noparitem \@donoparitem
  \else \if@inlabel \indent \par \fi
         \ifhmode \unskip\unskip \par \fi
         \if@newlist \if@nobreak \@nbitem \else
                        \addpenalty\@beginparpenalty
                        \addvspace\@topsep \addvspace{-\parskip}\fi
           \else \addpenalty\@itempenalty \addvspace\itemsep
          \fi
    \global\@inlabeltrue
\fi
\everypar{\global\@minipagefalse\global\@newlistfalse
          \if@inlabel\global\@inlabelfalse \hskip -\parindent \box\@labels
             \penalty\z@ \fi
          \everypar{}}\global\@nobreakfalse
\if@noitemarg \@noitemargfalse \if@nmbrlist \refstepcounter{\@listctr}\fi \fi
\setbox\@tempboxa\hbox{\makelabel{#1}}%
\global\setbox\@labels
 \hbox{\unhbox\@labels \hskip \itemindent
       \hskip -\labelwidth \hskip -\labelsep
       \ifdim \wd\@tempboxa >\labelwidth
                \box\@tempboxa
          \else \hbox to\labelwidth {\unhbox\@tempboxa}\fi
       \hskip \labelsep}\ignorespaces}

\newcounter{enumsi}

\def\@mklab#1{#1\hfil}
\def\@mklabr#1{\hfil#1}
\def\enummklab#1{(\eelabel)\hfil\hbox to1.25em{\hfil#1}}
\def\enummakelabel#1{\enummklab{#1}\global\let\makelabel=\@mklabr}

\def\eenumsentence{\@ifnextchar[{\@eenumsentence}
{\refstepcounter{enums}\@eenumsentence[\theenums]}}

\long\def\@eenumsentence[#1]#2{\def\eelabel{#1}%
\begin{list}{\alph{enumsi}.}{\usecounter{enumsi}%
\advance\leftmargin by1.75em\advance\labelwidth by1.75em%
 \setlength{\itemsep}{0pt} 
 \setlength{\parsep}{0pt} 
\let\makelabel=\enummakelabel}%
#2
\end{list}}

\def\rpar#1|#2{\param{#1}_{\mmsoa{#2}}}

\def\prpty#1|#2{\lbrack \, #1 \mid #2 \,\rbrack}

\def\lprpty#1|{\lbrack \, #1 \mid }

\def\chmmode#1{\ifmmode #1\else $#1$\fi}

\def\mmsoa#1{\chmmode{\langle \!\langle #1 \rangle \!\rangle}}

\def\param{\mathaccent"705F}
\def\mmrpar#1|#2{{\param{#1}}_{\mmsoa{#2}}}

 \def\numadvance{\global \advance \sentencenumber by 1}

 \def\param#1{#1}

  \newlength\titlebox 
 \twocolumn 

\def\addcontentsline#1#2#3{}

\def\maketitle{\par
 \begingroup
   \def\thefootnote{\fnsymbol{footnote}}
   \def\@makefnmark{\hbox to 0pt{$^{\@thefnmark}$\hss}}
   \twocolumn[\@maketitle] \@thanks
 \endgroup
 \setcounter{footnote}{0}
 \let\maketitle\relax \let\@maketitle\relax
 \gdef\@thanks{}\gdef\@author{}\gdef\@title{}\let\thanks\relax}
\def\@maketitle{\vbox to \titlebox{\hsize\textwidth
 \linewidth\hsize \vskip 0.625in minus 0.125in \centering
 {\vspace{-0.5in}\LARGE\bf \@title \par} \vskip 0.2in plus 1fil minus 0.1in
 {\def\and{\unskip\enspace{\rm and}\enspace}%
  \def\And{\end{tabular}\hss \egroup \hskip 1in plus 2fil
           \hbox to 0pt\bgroup\hss \begin{tabular}[t]{c}\Large\bf}%
  \def\AND{\end{tabular}\hss\egroup \hfil\hfil\egroup
	  \vskip 0.25in plus 1fil minus 0.125in
	   \hbox to \linewidth\bgroup\Large \hfil\hfil
 	     \hbox to 0pt\bgroup\hss \begin{tabular}[t]{c}\Large\bf}
  \hbox to \linewidth\bgroup\Large \hfil\hfil
    \hbox to 0pt\bgroup\hss \begin{tabular}[t]{c}\Large\bf\@author
			    \end{tabular}\hss\egroup
    \hfil\hfil\egroup}
  \vskip 0.3in plus 2fil minus 0.1in
}}

\skip\footins 9pt plus 4pt minus 2pt
\def\footnoterule{\kern-3pt \hrule width 5pc \kern 2.6pt }
\labelwidth\leftmargini\advance\labelwidth-\labelsep \labelsep 5pt

\def\@listi{\leftmargin\leftmargini}
\def\@listii{\leftmargin\leftmarginii
   \labelwidth\leftmarginii\advance\labelwidth-\labelsep
   \topsep 2pt plus 1pt minus 0.5pt
   \parsep 1pt plus 0.5pt minus 0.5pt
   \itemsep \parsep}
\def\@listiii{\leftmargin\leftmarginiii
    \labelwidth\leftmarginiii\advance\labelwidth-\labelsep
    \topsep 1pt plus 0.5pt minus 0.5pt
    \parsep \z@ \partopsep 0.5pt plus 0pt minus 0.5pt
    \itemsep \topsep}
\def\@listiv{\leftmargin\leftmarginiv
     \labelwidth\leftmarginiv\advance\labelwidth-\labelsep}
\def\@listv{\leftmargin\leftmarginv
     \labelwidth\leftmarginv\advance\labelwidth-\labelsep}
\def\@listvi{\leftmargin\leftmarginvi
     \labelwidth\leftmarginvi\advance\labelwidth-\labelsep}

\belowdisplayskip \abovedisplayskip
\def\@normalsize{\@setsize\normalsize{11pt}\xpt\@xpt}
\def\small{\@setsize\small{10pt}\ixpt\@ixpt}
\def\footnotesize{\@setsize\footnotesize{10pt}\ixpt\@ixpt}
\def\scriptsize{\@setsize\scriptsize{8pt}\viipt\@viipt}
\def\tiny{\@setsize\tiny{7pt}\vipt\@vipt}
\def\large{\@setsize\large{12pt}\xipt\@xipt}
\def\Large{\@setsize\Large{14pt}\xiipt\@xiipt}
\def\LARGE{\@setsize\LARGE{16pt}\xivpt\@xivpt}
\def\huge{\@setsize\huge{20pt}\xviipt\@xviipt}
\def\Huge{\@setsize\Huge{23pt}\xxpt\@xxpt}

\begin{document}

\title{A TOOL FOR COLLECTING DOMAIN DEPENDENT SORTAL CONSTRAINTS FROM CORPORA}
\author{
Fran\c{c}ois Andry\/$^*$, Mark Gawron, John Dowding,
and Robert Moore \\ \\
SRI International, Menlo Park, CA\\
\/$^*$CAP GEMINI Innovation, Boulogne Billancourt, France\\
Internet: andry@capsogeti.fr
}

\maketitle

\noindent
Topical paper : Tools for NL Understanding (Portability).\\

\section{ABSTRACT}
In this paper, we describe a tool designed to generate semi-automatically
the sortal constraints specific to a domain to be used in a natural
language (NL) understanding system. This tool is evaluated using the SRI
Gemini NL understanding system in the ATIS domain.\\

\section{INTRODUCTION}

The construction of a knowledge base related to a specific
domain for a NL understanding system is time
consuming. In the Gemini system, the domain-specific
knowledge base includes
a {\it sort hierarchy} and a set of {\it sort rules}
that provide (largely domain-specific) selectional
restrictions for every predicate invoked
by the lexicon and the grammar.
The selectional restrictions
provide a source of constraints over
and above syntactic constraints
for choosing the correct analysis of a sentence.
The sort rules are
generally entered by a linguist, by hand, from the study of a
corpus and while tuning the grammar.

However, the use of an interactive tool that can help the linguist to
acquire this knowledge from a corpus\cite{GRI86}\cite{LAN88}, can
drastically reduce the time dedicated to this task, and also
improve the quality of the knowledge base in terms of both accuracy
and completeness. The reduction in the amount of effort to develop the
knowledge base becomes obvious when porting an existing system to a
new domain. At SRI, our main concern was to port Gemini, our NL understanding
system to other domains
without investing the same amount of work we put into the first
domain application\footnote{The actual domain is
Air Transportation (ATIS) used as a benchmark in the ARPA community.}.

In this paper, we describe the results of using this
semi-automatic tool to port the Gemini NL system to the
ATIS domain, a domain that Gemini had already been ported to, and
for which it had achieved high performance and grammatical coverage using
hand-written sortal
constraints.  Choosing a known domain, rather than a new one,
allowed us to compare the performance of the derived sorts to the hand-written
ones, holding the domain, grammar, and lexicon constant.
It also allowed us to evaluate the semi-automatically
obtained coverage using the evaluation tools provided for
the ATIS corpus.\\

\section{PARSING WITH SORTS}

Gemini\cite{DOW93} implements a clear separation between
syntactic and semantic information. Each syntactic
node invokes a set of semantic rules
which result in the building of a set of logical
forms for that node.
Selectional restrictions are enforced on the logical forms
through the sorts mechanism: All predications
in a candidate logical form must be licensed by some
sorts rule. The sorts are located in a conceptual hierarchy of approximately
200
concepts and are implemented as
Prolog terms such that more general sorts subsume more specific
sorts\cite{MEL88}. Failure to match any available
sorts rule can thus be implemented as
unification-failure.

Gemini parser creates logical forms expressions like the following one :

\noindent
$exists((A;[flight]), $\\
\hspace*{3mm}$[and, [flight, (A;[flight])];[prop],$\\
\hspace*{6mm}$[to, (A;[flight]),$\\
\hspace*{12mm}$('BOSTON';[city])];[prop]];[prop]);[prop]$

In these logical form expressions, every sub-expression is
assigned a sort, represented as the right-hand-side of a '{\bf ;}'
operator\cite{ALS92}.
Sorts rules for predicates
are declared with {\tt sor/2} clauses:

\begin{center}
${\tt sor}('BOSTON', [city]).$
${\tt sor}(to, ([[flight],[city]], [prop])).$
\end{center}

The above declarations license the use of 'BOSTON'
as a zero-arity predicate
with ``resulting'' sort $[city]$
and 'to' as a two-place predicate relating flights
and cities with resulting sort $[prop]$ (or proposition).

In the ATIS application domain, for example, the subject (or
actor) of the
verb {\it depart}, as in '{\it the morning flights departing for denver}',
can be a flight. For this, we use the following set of sort
definitions:

\begin{center}
$sor(depart,([[departure]],[prop]))$ \\
$sor(flight,([[flight]],[prop]))$ \\
$sor(actor,([[departure],[flight]],[prop]))$ \\
\end{center}
\noindent
The first two definitions make $depart$ and $flight$
predicates compatible with departure and flight events
respectively, returning
a proposition;
the third makes $actor$ a relation
that can hold between departure and flight,
also returning a proposition.
A simple example of a logical form licensed
by these rules follows (with the result sort $[prop]$ suppressed):

\small
\noindent
$qterm(some,((X;[flight]),$\\
\hspace*{3mm}$[and,[flight,(X;[flight])],$\\
\hspace*{7mm}$exists((Y;[flight]),$\\
\hspace*{11mm}$[and, [depart,(Y;[departure])],$\\
\hspace*{20mm}$(Y;[departure]),$\\
\hspace*{20mm}$[actor,(Y;[departure]),(X;[flight])]])])$
\normalsize

Which would be roughly the logical form for '{\it a departing flight}'.

\section{SORT ACQUISITION}

\begin{figure}
\setlength{\unitlength}{0.1in}
\begin{picture}(30,35)(0,0)
\put(13,32){\framebox(10,3){Corpus}}
\put(18,32){\vector(0,-1){5}}
\put(18,25){\oval(8,4)}
\put(15.7,24.5){Parser}
\put(0,23.5){\framebox(9,3){Signature File}}
\put(9,25){\vector(1,0){5}}
\put(9.3,25.4){step=1}
\put(18,23){\vector(0,-1){5}}
\put(13,15){\framebox(10,3){Sorts + Stats}}
\put(18,15){\vector(0,-1){4}}
\put(18,9){\oval(8,4)}
\put(16,9.5){Editor}
\put(16.5,8){Tool}
\put(18,7){\vector(0,-1){4}}
\put(13,0){\framebox(10,3){New Sort File}}
\put(23,1.5){\line(1,0){5}}
\put(28,1.5){\line(0,1){23.5}}
\put(28,25){\vector(-1,0){6}}
\put(22.5,25.4){steps $>$ 1}
\end{picture}
\caption{Iterative Acquisition of Sorts.}
\label{fig:acquisition}
\end{figure}
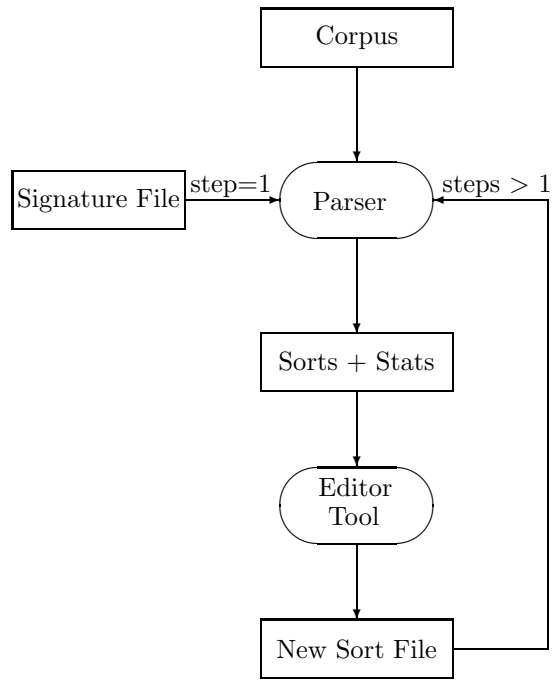

The approach we have taken is to
start from an initial ``schematic'' sorts file
we call the signature file (explained below),
which essentially allows all predicate
argument combinations.
We then harvest a set of preliminary sort rules by
parsing a large corpus.
The logical forms that induce these preliminary rules come from
parses that essentially incorporate
only syntactic constraints. The resulting sorts
rules are filtered by hand and
the process is iterated
with an increasingly accurate
sorts file, converging rapidly on the sorts file specific
to the application domain (fig.~\ref{fig:acquisition}).

\subsection{Signature and Restrictions}

If we started the above iteration process
with {\it no} sortal information,
then the logical forms resulting
from a parse would contain no sortal information,
and only vacuous sortal rules would
be harvested.

The first step is thus to build an initial sort file we call the {\it
signature} file. The idea is to assign lexical predicates
inherent sorts, but not to
assign any rules which constrain which
lexical items can combine with which. The
signature file, then, is not just domain-independent. It has no
information at all about semantic combinatorial possibilities, not
even those determined by the language (for example, that the verb {\it
break} does not allow propositional subjects). The reason for this is
so that it can be generated largely automatically from the lexicon.

\subsection{The Signature}

Lets begin with certain inherently relational predicates,
for which the signature file gives only
an arity and the result sort. For example the signature for the
predicates {\it at} (corresponding to the preposition)
and {\it actor} (corresponding to logical subject) would be
the same:

\begin{center}
$signature(at,([X,Y],[prop])$\\
$signature(actor,([X,Y],[prop])$\\
\end{center}

This signature is used as the sort rule for
{\it at} and {\it actor} in the sorts tool's first iteration.
The effect is to limit the choice of sorts rules for
these predicates to rules which are further instantiations
their signatures,
that is, to rules licensing them to take two arguments of
any sort to make a proposition. The object
in successive iterations will be to assign
these relational predicates substantive
sortal constraints,
thus constraining head modifier relations
and the parse possibilities.

Verbs, nouns, some adjective and adverbs,
on the other hand, have
signatures with fully or partially instanciated arguments:
For example, in the ATIS
domain, the verbs {\it depart}, {\it get\_in}
and the nouns {\it data}, {\it flight} have the signatures:

\noindent
$signature(depart,([[departure]],[prop]))$\\
$signature(get\_in,([[arrival]],[prop]))$\\
$signature(data,([[information]],[prop]))$\\
$signature(flight,([[flight]],[prop]))$ \\

\noindent
These declarations have no effect on
the combinatorial possibilities of these words
(they tell us nothing about what can be the subject
of the verb {\it depart} or what verbs
the noun {\it flight} can be subject of),
but when a logical form is built up from
a syntactically licensed parse (like
the one given above for
{\it a departing flight}), these sortal declarations
will ``fill in'' the sorts for the
connecting predicate {\it actor}, generating
the sort rule:
\begin{center}
$signature(actor,([[departure],[flight]],[prop])$\\
\end{center}
Thus in the signature file, lexical predicates have their
own ``inherent'' sort rules,
which then help build up the
sort rules for the relational predicates.
The inherent sort rules
for adjectives like {\it cheap} and {\it late} will
constrain only their first argument. The reason for this
is that it is this first argument that
modifiers (such as intensifying adverbs and specifiers), will
hook on to.\\

\noindent
$signature(cheap,([[cost\_soa],A,B],[prop]))$\\
$signature(late,([[temporal\_stage],A,B],[prop]))$\\

\noindent
At the same time the argument position
filled in by what the adjectives modify
is left unconstrained. The signature
file thus makes no commitment about what sorts
of things can be {\it late} or
{\it cheap}; it just needs to say there is
such a thing as lateness and cheapness.
This is why for a new domain  the signature
file can be generated largely automatically,
using a new inherent sort for each
new lexical item, assigning the
type of predicate appropriate to its grammatical
category.

All zero-arity predicates (names) need to have
inherent sorts.  Certain
general 'tool words' which include numbers, dates, time,
and commons words, will receive the same signatures
in any domain :

\noindent
$signature(3,([number]))$\\
$signature(friday,([[day]],[prop]))$\\
$signature(pm,([nonagent]))$\\
$signature(yes,([prop]))$\\

\noindent
In addition to this, however,
there is a whole list of words
specific to the domain which
need to be inherently sorted.
This part of
creating a signature file will need
to be done by hand:

\noindent
$signature('NASHVILLE',([city]))$\\
$signature('AIR\_CANADA',([airline]))$\\
$signature('LA\_GUARDIA',([airport]))$\\

\subsection{Extracting the Sorts}

We now give a more detailed example of
how sort rules are extracted from logical
forms (LFs) built by the parser.
For '{\it the morning flights flying to denver}',
we obtain roughly the following Logical Form :

\noindent
$qterm(the;[non\_symmetric\_determiner],$\\
\hspace*{4mm}$A;[flight],$\\
\hspace*{8mm}$[and,$\\
\hspace*{10mm}$[flight,(A;[flight])],$\\
\hspace*{10mm}$[n\_n\_rel,$\\
\hspace*{14mm}$(B;[day\_part])$ \ $[and,$\\
\hspace*{30mm}$[morning,$\\
\hspace*{30mm}$(B;[day\_part])]]$\\
\hspace*{12mm}$;[[day\_part]],[prop],$\\
\hspace*{12mm}$A;[flight]],$\\
\hspace*{10mm}$exists(C;[flight],$\\
\hspace*{24mm}$[and,$\\
\hspace*{26mm}$[fly,(C;[flight])],$\\
\hspace*{26mm}$[actor,(C;[flight]),$\\
\hspace*{37mm}$(A;[flight])],$\\
\hspace*{26mm}$[has\_aspect,$\\
\hspace*{34mm}$(C;[flight]),$\\
\hspace*{34mm}$(in\_progress;[aspect])],$\\
\hspace*{26mm}$[to,(C;[flight]),$\\
\hspace*{34mm}$('DENVER';[city])]])])$\\
$;[flight]$\\

The extraction process consists of a recursive exploration of the
logical form and retrieval of each predicate and its arguments. For
example, from the LFs above, our
tool would extract the following sort definitions
set\footnote{For reason of efficiency and simplification, we exclude
some very common predicates independent of the domain,
such as '{\it and}', '{\it equal}', '{\it exists}',
'{\it has\_aspect}', and '{\it qterm}'. } :

\small
\noindent
$sor(flight,[[flight]],[prop])$\\
$sor(morning,[[day\_part]],[prop])$\\
$sor(n\_n\_rel,[([[day\_part]],[prop]),[flight]],[prop])$\\
$sor(fly,[[flight]],[prop])$\\
$sor(actor,[[flight],[flight]],[prop])$\\
$sor(to,[[flight],[city]],[prop])$\\
$sor(frag\_np,[[flight]],[prop])$\\
$sor(np\_frag,[[prop]],[prop])$\\
\normalsize

When constrained only by signatures,
the parser typically finds a
large number of logical forms.
The sorts tool provides the option of
harvesting sort rules in one of
two ways,
either from all generated logical
forms, or only from the Preferred Logical Form (PLF).
The parse preference component implemented
in Gemini
chooses the best intepretation from the chart,
based on syntactic heuristics\cite{DOW93}, and provides a set of
PLFs.

In addition to the extraction of the sort rules, we also calculate
the occurrence $\Theta_{i}$ of each sort rule for all the sentences
of the corpus. We then normalized $\Theta_{i}$  by the number of
logical forms that include the sort rule (${\overline \Theta_{i}}$).
Each value ${\overline \Theta_{i}}$ is stored along with its sort
rule and used to calculate the probabilities related to the sort
rule :

\begin{center}
$Prob(Sort_{i}) = ${\LARGE
${\frac{{\overline\Theta_{i}}}{\sum_{i=0}^{n}{\overline \Theta_{i}}}}$}\\
\end{center}


In fact three sets of probabilitilies are calculated for each rule R:
(1) Global probability of sort rule R: the number of invocations
of rule R normalized by the number of LFs containing R and
divided by the total number of rule invocations
in the corpus; (2) Conditional probability of  rule R
given a particular predicate; (3) Conditional probability of R
given the predicate in R and an argument of the same sort as
the first argument of R.

Also, associated to each sort definition, we keep
the list of the indexes of a small set of sentences which contain the
corresponding sort definition in its logical form. This set is used as
a sample for the set editor tool.\\

\subsection{The Argument Restrictions}

The argument restrictions
are instantiated versions of the signatures for each
predicate. For example, after parsing and extraction from
the logical forms, the arguments $X$ and $Y$ of the signature
associated to the preposition {\it at} will help to generate
a list of several sort definitions such as :\\

\noindent
$sor(at,([[airport],[city]],[prop])$\\
as in : '{\it the aiport at Dallas}',\\

\noindent
$sor(at,([[domain\_event],[time\_point]],[prop])$\\
as in : '{\it departure at 9pm}'.\\

\section{SORT EDITING}

At each step of the process, after parsing, the linguist, using
the interactive sort editor, can examine the new sort file which
has been generated and choose which sortal definition need to be
eliminated. Statistical information associated to each sort
definition helps him decide which ones are revelant or not.
We have also included the possiblility of adding a sort definition,
although this kind of operations should be very rare.
In fact the main activity of the linguist
using the sort editor tool, will be to filter the
sort definitions generated by the parsing of the corpus.

\subsection{Description of the tool}

The sort editor tool is an interactive, window-based program.
It has a main window for displaying and editing the sorts
and a set of buttons that help the user to either display
additional information or perform actions such as :

\begin{itemize}
\item{load or save a sort file,}
\item{select a functor among the list of all functors
and display the list of its possible arguments, result and
probabilities,}
\item{deletion and insertion of a sort definition,}
\item{display a sample of sentences associated to a specific sort definition,}
\item{mapping between the sort definitions and a reference sort file (for
evaluation),}
\item{changing the way the sort definitions are displayed (result or not,
mapping or not, global probability, conditional to a functor, or relative to
the first
argument of a definition),}
\item{use of a threshold on the probabilities to filter the sort definitions,}
\item{retrieve the list of functors given a certain argument,}
\item{display the sentences associated to a sort definition,}
\item{display the list of predicates which have been excluded form the
extraction,}
\item{specification of a sortal hierarchy to be used with the sort
definitions for the next iteration,}
\item{use of a whiteboard to save specific sentences and information during a
session.}
\end{itemize}

The tool uses ProXT, the Quintus Prolog interface to MOTIF widget set
and the X-Toolkit.\\

\section{EVALUATION AND RESULTS}

Evaluate the porting to a new domain require measuring
how the new sort file contributes to
perform the target task within the new domain.
This kind of evaluation is difficult because it is
hard to separate the contribution of the grammar
and the contribution of the sorts constraints. One
way to evaluate our tool would be to have a
file of `` correct'' sortal constraints
that we use as a reference to check
the ones we generate with our tool. The problem is that this kind
of file does not exist for new domains, since obtaining
such file is precisely the purpose of our tool.

The
approach we have chosen was to use the sort file built by hand for
the ATIS corpus and to check this 'reference file' against the new
sort file we intend to build, using our tool on a corpus of the
same domaine.

\subsection{Building the signature file}

For the this first experimental
exercise with the sort tool,
we built the signature file somewhat
differently than we would build it
for a new application. In order
to facilitate evaluating the tool, our
goal this time was to come up with
a signature file
be compatible with the reference
file built by hand.

The first step in the experiment
was to automatically extract the signatures from
the lexicon and reference sorts file, which contains nearly 2200 sort
definitions.
Signatures are largely predictable from the grammatical category of a word
For example, most of the verbs (except the
auxiliaries) with one argument, received a signature
identical to the sort definition. On the other hand, most of
the prepositions received a signature with all their
arguments replaced by a variable (since
they are domain-specific).
In this maiden voyage
of the sort acquisition system, the
signatures chosen for verbs, adjectives and nouns
were made compatible with the sort hierarchy
used by the reference sorts file.
In porting to a new domain, the lexical
signatures would presumably
use an automatically generated sort hierarchy,
almost entirely flat, with a unique lexical sort
for each lexical item.

In addition to this, some signatures,
for logical predicates and predicates
introduced in semantic rules,
were added by hand. These represent a little bit more
than $15\%$ of the final
signature file which contains a total of 1357 signatures.
Half of these signatures are
zero-arity predicates mostly automatically built
from the lexicon.

\subsection{Parsing Madcow}

The next step of our experiment was to parse a corpus
from the ATIS domain
using the signature file we have built. For this, we have
used the MADCOW corpus\cite{HIR92}, that includes 7243
sentences of various length (from 1 to 36 words) with a
large linguistic coverage from this domain. This process
had been done in both modes LFs and PLFs. The idea was to
compare the result in both modes, to check whether
the use of parsing preferences was relevant for the
extraction of the sort definitions or if we had to use
all the Logical Forms from the parsing.

The first iteration of parsing MADCOW produced 5917 and 2275 sort
rules\footnote{Since zero-arity sort predicates have a signature
identical to their sort rule, only sorts rules with at least an
argument were extracted during the parsing of MADCOW.}
respectively for the LFs and PLFs modes.

\subsection{Mapping corpus and reference rules}

For this first evaluation, we also used a feature of our tool
which can map each sort rule produced
by the extraction phase against the rules of a reference
sort file. The mapping consists of assigning one of the
following categories to each corpus acquired sort rule :

\begin{itemize}
\item{Exact : the corpus rule match exactly with a reference rule,}
\item{Incompatible : the corpus rule does not match with any reference rule,}
\item{Subsumed-by : the corpus rule is subsumed by at least one
reference rule,}
\item{Subsumes : the corpus rule subsumes at least one reference rule,}
\item{Incomparable : the corpus rule is incomparable\footnote{Two sort
rules are incomparable, when they unify each other while none of them
subsumes the other one.} with at least one reference rule.}
\end{itemize}

The following table shows the repartition of mapping categories
modes LFs and PLFs :

\begin{center}
\begin{tabular}{| l | r |  r|}
\hline \hline
 & {\em LFs} & {\em PLFs } \\
\hline
Exact & 409 & 362 \\
\hline
Incompatible & 3055 & 691 \\
\hline
Subsumed-by & 1557 & 888 \\
\hline
Subsumes & 375 & 156 \\
\hline
Incomparable & 521 & 178 \\
\hline
\hline
Total & 5917 & 2275 \\
\hline \hline
\end{tabular}
\end{center}

The first comments concerning these figures is that the percentage of
incompatible rules is higher for the LFs than the PLFs mode
(respectively $52\%$ vs $30\%$), and the number
of 'exact' sorts is more than half for LFs than PLFs. This shows
that the use of Preferred Logical Forms for parsing is more efficient
in extracting  the 'good sorts'.

However, the figures do not give an exact idea of the completeness
and precision of our tool,
since there is a large number of rules
subsumed by other ones (more than $30\%$ for LFs and almost
$50\%$ for PLFs mode). In fact, some of the corpus rules are
subsumed by more general rules in the reference sort file
while providing the same coverage as the reference sort rules.

Therefore, the {\bf precision} of our tool for the
PLFs mode just after the extraction phase can be estimated
between $16\%$ (exacts rules)
and $55\%$ (exact rules plus subsumed rules). This number
gets better and more precise very quickly after the first
iteration of editing since the work of the linguist is
precisely to remove most of the incompatible and incomparable
rules and rules which are either too general or too specific.

The {\bf overgeneration} of the tool just
after parsing, for the PLFs mode, can be estimated to at
least $30\%$ (the percentage of incorrect rules). After
the first iteration of editing, this number decreases
very quickly since low probabilities help the linguist
to eliminate rules that are incompatible or
incomparable.

The {\bf recall} for the PLFs mode after parsing,
which is the ratio of the
'Exact' corpus rules by the number of reference rules used for
the mapping in our evaluation (636 non zero-arity sorts rules),
can be estimated to at least $57\%$.

A more precise estimation of the exact number of 'Exact'
rules could be computed by using the sortal hierarchy, and
generate for the two sets of rules (corpus and reference)
all the rules that can be subsumed, and realize the
mapping only with these rules. \\

\section{CONCLUSION}

This first evaluation of our tool in the ATIS domain
shows that the acquisition of sorts from a corpus can be
partially automated, reducing drastically the time the
linguistic dedicates to this task (the precision
converges in few editing iteration). In addition
to this, the possibility of a systematic examination for
all predicates with crosschecking tools such as sentence
visualisation and functor browing helps the linguist to
establish strict aquisition methods for the knowledge
base in new domains.

In addition to this, the tool can also be used to
improve an existing knowledge base. For example,
the study of the incompatible rules during
this first evaluation helped us to discover
new rules that will increase the coverage of
Gemini in the ATIS system.\\

\section{Acknowledgements}

{\small
This research was supported by the
Advanced Research Projects Agency under contract
with the Office of Naval Research, and
by a grant of the Lavoisier Program from the French
Foreign Office. The views and conclusions contained in this
document are those of the authors and should not be interpreted
as necessarily representing the official policies, either expressed
or implied, of the Advanced  Research Projects Agency of the U.S.
Government, or those of the Scientific Mission of the French Foreign
Office.
}\\


\begin{thebibliography}{99}

\bibitem{ALS92} Alshawi, H. (ed.),
{\it The Core Language Engine}, MIT Press, 1992.

\bibitem{DOW93}
Dowding J., Gawron J.M., Appelt D., Bear J., Cherny L.,
Moore R. and Moran D.,
``GEMINI : A Natural Language System For Spoken-Language
Understanding'', Proceedings of the 31st Meeting of the
Association for Computational Linguistics, Ohio State University,
Columbus, Ohio, pp. 54-61, 1993

\bibitem{GRI86} Grishman R., Hirschman L. and Ngo T.N.,
``Discovery Procedures for Sublanguage Selectional
Patterns : Initial Experiments'',
{\it Computational Linguistics}, Vol. 12:3 pp. 205, 1986.

\bibitem{HIR92} Hirschman L.,
``Multi-Site Data Collection for a Spoken Language Corpus'',
MADCOW, in {\it Proceedings of the DARPA Speech and Natural
Language Workshop}, pp. 7-14, Feb. 1992.

\bibitem{LAN88} Lang F.M., Hirschman L.,
``Improved Portability and Parsing Through Interactive
Acquisition of Semantic Information'',
In {\it Second Conference on Applied Natural Language
Processing'}, Feb. 1988.

\bibitem{MEL88} Mellish, C., ``Implementing Systemic
Classification by Unification''.  {\it Computational Linguistics},
Vol. 14, pp. 40-51, 1988.

\end{thebibliography}
\end{document}